\documentclass[11pt]{article}

\usepackage{agenticlearning-xelatex}
\hypersetup{pdftitle={The Surprising Effectiveness of Deleting Weights in LLM Post-Training}, pdfauthor={Jack Lu, Zhenbang Yang, Mike Lasby, Tejas Pote, Yani Ioannou, Mengye Ren}} %
\usepackage{mathtools}
\usepackage{float}
\definecolor{KeptBlue}{HTML}{185FA5} %
\definecolor{DeletedRed}{HTML}{A32D2D} %
\newcommand{\ourmethod}{Deletion Tuning}
\newcommand{\acronym}{Deletion Tuning}
\newcommand{\ourmethodit}{\textit{\ourmethod{}}}
\newcommand{\acronymit}{\acronym{}}

\makeatletter
\newcommand{\blfootnote}[1]{%
  \begingroup
  \renewcommand\@makefntext[1]{##1}
  \footnotetext{#1}%
  \endgroup
}
\makeatother

\shorttitle{\looseness=-10000 The Surprising Effectiveness of Deleting Weights in LLM Post-Training}
\shortauthor{\looseness=-10000 Lu \etal}

\newcommand{\MR}[1]{} %
\newcommand{\JL}[1]{} %

\begin{document}
\title{The Surprising Effectiveness of Deleting Weights in\\LLM Post-Training}
\providecommand{\iflatexml}{\iffalse}
\iflatexml
\author{Jack Lu\textsuperscript{*} \\ New York University \and Zhenbang Yang\textsuperscript{*} \\ New York University \and Mike Lasby \\ University of Calgary \and Tejas Pote \\ University of Calgary \and Yani Ioannou \\ University of Calgary \and Mengye Ren \\ New York University}
\else
\author{
  Jack Lu$^{*1}$, Zhenbang Yang$^{*1}$, Mike Lasby$^2$, Tejas Pote$^2$, Yani Ioannou$^2$, Mengye Ren$^1$\\
  $^1$New York University, $^2$University of Calgary
  \\
  \url{https://agenticlearning.ai/deleting-weights/}
}
\fi
\date{}

\iflatexml\else\blfootnote{* Equal contribution.}\fi

\maketitle
\iflatexml
\begin{center}* Equal contribution. Project page: \url{https://agenticlearning.ai/deleting-weights/}\end{center}
\fi

\begin{abstract}
\looseness=-10000
How little of a pretrained large language model (LLM) has to change for it to acquire new capabilities?
We find that zeroing fewer than $0.05\%$ of its weights, with no other modification, matches full fine-tuning (FFT) at scale across three LLM post-training regimes: reinforcement learning with verifiable rewards, supervised fine-tuning, and on-policy distillation.
Our method, \ourmethodit{}, learns a binary mask over the gate projection of transformer blocks.
Two practical advantages follow: the trained mask is hundreds to thousands of times smaller than the corresponding FFT checkpoint, and \acronymit{} forgets far less, which addresses a central challenge in continual learning. Under reinforcement learning, its prior-task accuracy stays near the base level while FFT, LoRA, and random sparse fine-tuning lose up to $5.4$ points.
To understand why masking alone suffices, we analyze the geometry of the learned weight changes under reinforcement learning using stable rank, spectrum drift, and principal-weight overlap.
Compared with LoRA and random sparse fine-tuning, \acronymit{} is closest to FFT across the three measures.
\end{abstract}

\section{Introduction}
\label{sec:intro}

\looseness=-10000
To acquire new capabilities during LLM post-training, full fine-tuning can alter every parameter, LoRA~\citep{lora} learns a low-rank update, and the Multiple Ticket Hypothesis (MTH)~\citep{mth} and SHiRA~\citep{shira} optimize values at selected weight positions.
Weight deletion is typically used differently: pruning removes weights while preserving an already-trained model's capabilities for efficient inference~\citep{wanda, sparsegpt}, and the Lottery Ticket Hypothesis finds sparse subnetworks that can be trained to match dense networks~\citep{lotteryticket}.
Unlike full fine-tuning, LoRA, and sparse updates, deletion writes no new values and only chooses which pretrained weights to keep. Recent work shows that fine-tuned solutions lie close to the pretrained model. RL fine-tuning changes only a small subnetwork~\citep{rlsmallsubnet}, and task experts lie densely around pretrained weights~\citep{neuralthickets}. We push this observation to its limit across reinforcement learning with verifiable rewards (RLVR), supervised fine-tuning (SFT), and on-policy distillation (OPD), and ask:
\begin{center}
\emph{Can a pretrained LLM acquire new capabilities by learning only which weights to set to zero?}
\end{center}

\looseness=-10000
The answer is \emph{yes} across all three post-training regimes.
We propose \ourmethodit{}, which learns only a binary mask over the gate-projection weights of transformer blocks.
At deployment, each target weight either retains its pretrained value or is set to zero, so the learned binary mask is the entire adapter.
We test \acronymit{} with Group Relative Policy Optimization (GRPO)~\citep{grpo} for RLVR, standard SFT, and Self-Distillation Fine-Tuning (SDFT)~\citep{sdft} for OPD.
Our experiments span two model families, eight backbones from $0.5$B to $30$B, and both in-distribution and out-of-distribution evaluation.
At the largest model scale tested in each setting, \acronymit{} matches full fine-tuning (FFT) across reasoning, domain knowledge, and tool-use tasks.

\looseness=-10000
The binary-mask form brings two practical advantages: (i) \emph{the adapter is compact}, with the trained mask hundreds to thousands of times smaller than the corresponding FFT checkpoint and smaller than even LoRA $r{=}1$ in our main GRPO comparison; and (ii) \emph{\acronymit{} forgets less}, with its forward KL remaining low and its prior-task accuracy staying near the base level across all three Llama scales. Under FFT, LoRA, and random sparse fine-tuning, forward KL grows and prior-task accuracy falls (Figure~\ref{fig:retention}), consistent with forward KL as a predictor of forgetting~\citep{rlrazor}. Learning new tasks while keeping old ones makes \acronymit{} a promising candidate for continual learning.

\looseness=-10000
Finally, we examine why \acronymit{} can match FFT with deletion alone.
We confirm that the gains come from learning the mask rather than the Wanda prior used in its initialization.
We also find that the geometry of \acronymit{}'s learned weight updates tracks FFT's more closely than LoRA or random sparse fine-tuning across stable rank, spectrum drift, and principal-weight overlap.

\looseness=-10000
Our contributions are as follows:
\begin{itemize}
\item \looseness=-10000
\textbf{Deletion alone matches FFT.} \acronymit{} learns which gate-projection weights to delete (Section~\ref{sec:method}). It is on par with FFT at the largest scale we test in GRPO, SFT, and SDFT, up to a 30B mixture-of-experts model (Sections~\ref{sec:results} and~\ref{sec:results-sft-sdft}). Under GRPO, it also solves problems that the base model fails in all $128$ sampled attempts (Section~\ref{sec:results}).
\item \looseness=-10000
\textbf{\acronymit{} is compact and forgets less.}
Its masks are orders of magnitude smaller than FFT checkpoints in all three regimes (Sections~\ref{sec:results} and~\ref{sec:results-sft-sdft}), and it keeps prior-task accuracy throughout training (Section~\ref{sec:retention}).
\item \looseness=-10000
\textbf{\acronymit{}'s learned updates have FFT-like geometry.} Its weight updates resemble FFT's on three geometric measures (Section~\ref{sec:geometry}), which helps explain its accuracy.
\end{itemize}

\section{Related Work}
\label{sec:related}

\paragraph{LLM post-training.}
\looseness=-10000
We study three post-training regimes that differ in where the training signal comes from. Supervised fine-tuning (SFT) imitates fixed labeled targets~\citep{flan, instructgpt, alpaca}. Reinforcement learning optimizes a reward on the model's own samples, from PPO-based RLHF~\citep{ppo, instructgpt} to GRPO~\citep{grpo} with verifiable rewards, the recipe behind DeepSeek-R1~\citep{deepseekr1}. With verifiable rewards, rule-based checkers or LLM verifiers score each sample~\citep{verifierpayoff}. DPO~\citep{dpo} instead turns preferences into a closed-form loss. On-policy distillation trains a student on its own rollouts to match a teacher's per-token distribution~\citep{gkd, sdft, opsd}, and SDFT uses it to reduce catastrophic forgetting~\citep{ewc}. Self-distillation can also teach reasoning behaviors~\citep{skillfactory}. We use all three to test whether deletion works as a general post-training mechanism.

\paragraph{Sparsity and selective parameter updates.}
\looseness=-10000
One-shot LLM pruning methods like Wanda~\citep{wanda} and SparseGPT~\citep{sparsegpt} extend classical magnitude-based pruning~\citep{han2015pruning, zhu2017prune} to LLM scale, while continuous relaxations~\citep{l0reg, concrete, gumbelsoftmax} make binary masks differentiable. Unstructured pruning can even retain the test-time scaling of reasoning models~\citep{monjur2026pruning}. Earlier mask learning adapted vision backbones and encoder-only language models to downstream tasks such as image and sentence classification~\citep{piggyback,zhao2020masking,radiyadixit2020fine}.
Recent mask methods target semi-structured LLM inference sparsity~\citep{maskllm}, refinement of a fine-tuned LLM~\citep{mft}, or vision-language backbones~\citep{ramt}. \acronymit{} instead makes the mask the entire post-training update of a pretrained LLM across RLVR, SFT, and OPD, and it learns how many weights to delete.
Sparse weight tuning writes float deltas at selected positions~\citep{diffpruning, fishmask, mth, lift, shira}, additive adapters introduce trainable float parameters~\citep{lora, tinylora, transformer2, houlsby_adapters, prompttuning, contexttuning}, and robust fine-tuning methods restrict or regularize movement of pretrained weights under distribution shift~\citep{surgical_finetuning, spd}. \acronymit{} instead keeps every retained weight at its pretrained value and learns only which gate-projection weights to delete.

\paragraph{Understanding LLM fine-tuning.}
\looseness=-10000
The classical Lottery Ticket Hypothesis~\citep{lotteryticket} posits that randomly initialized networks contain sparse subnetworks that train to comparable accuracy in isolation. Recent work characterizes pretrained-model fine-tuning along three axes. \emph{Density:} task-expert solutions densely populate the neighborhood of pretrained weights~\citep{neuralthickets}. \emph{Geometry:}~\citet{rlsmallsubnet} report that RL fine-tuning naturally updates only $5$ to $30\%$ of parameters across PPO, GRPO, DPO, and several model families. \citet{pathnottaken} show these updates land off the principal weights of $W_0$ while leaving the spectrum largely intact, where principal weights are the top-magnitude entries of the rank-$k$ SVD reconstruction of $W_0$~\citep{lift}. \emph{Forgetting:}~\citet{rlrazor} establish that the forward KL between fine-tuned and base policies on the new task accurately predicts forgetting. \acronymit{} reproduces the off-principal update pattern and low spectrum drift. \acronymit{}'s low forward KL also predicts the strong prior-task retention we measure (Section~\ref{sec:retention}).

\begin{figure}[t]
\centering
\includegraphics[width=1.0\linewidth]{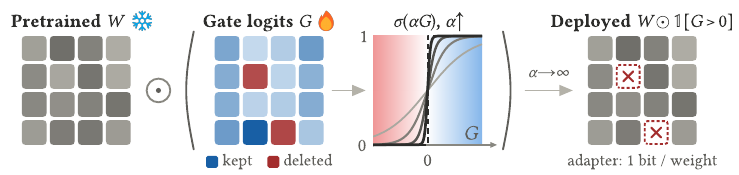}
\caption{
\looseness=-10000
\textbf{\ourmethodit{} learns which pretrained weights to delete and writes no new values.} The pretrained weights $W$ stay frozen, and each is paired with a trainable logit in $G$, the only parameters that post-training updates. During training, each weight is multiplied by $\sigma(\alpha G)$, a smooth mask between $0$ and $1$. The sharpness $\alpha$ rises over training until the sigmoid becomes a hard threshold at $G = 0$. At deployment, weights with positive logits (\textcolor{KeptBlue}{blue}) keep their pretrained values and weights with negative logits (\textcolor{DeletedRed}{red}) are set to zero (crossed out), so the trained adapter is one bit per weight.
}
\label{fig:method}
\end{figure}

\section{Background}
\label{sec:background}

\looseness=-10000
Our experiments use three post-training objectives.

\paragraph{Supervised fine-tuning (SFT).}
\looseness=-10000
Given a dataset $\mathcal{D}$ of prompts $x$ and target responses $y$ of length $L = |y|$, SFT minimizes the negative log-likelihood of the target tokens, which are fixed in advance and never sampled from the model:
\begin{equation}
\mathcal{L}_\text{SFT}(\theta) \;=\; -\,\mathbb{E}_{(x,y) \sim \mathcal{D}} \sum_{j=1}^{L} \log \pi_\theta(y_j \mid x, y_{<j}).
\label{eq:sft}
\end{equation}

\paragraph{Group Relative Policy Optimization (GRPO).}
\looseness=-10000
For RLVR, we use GRPO~\citep{grpo}. For each prompt $x$, the policy before the update, $\pi_{\theta_\text{old}}$, samples a group of $n$ responses $\{y^i\}_{i=1}^{n}$, and a verifier assigns each one a reward $r^i = r(x, y^i)$. Rather than learning a value function as in PPO~\citep{ppo}, GRPO compares each response with the rest of its group, $A^i = (r^i - \mathrm{mean}(r))/\mathrm{std}(r)$, and gives this advantage to every token of $y^i$ in PPO's clipped objective,
\begin{equation}
\mathcal{L}_\text{GRPO}(\theta) \;=\; -\,\mathbb{E}_{x,\,\{y^i\}}\,\frac{1}{n}\sum_{i=1}^{n}\frac{1}{|y^i|}\sum_{j=1}^{|y^i|}\min\!\Bigl(\rho^i_j A^i,\;\mathrm{clip}\bigl(\rho^i_j,\,1{-}\epsilon,\,1{+}\epsilon\bigr)\,A^i\Bigr),
\label{eq:grpo}
\end{equation}
\looseness=-10000
where $\rho^i_j = \pi_\theta(y^i_j \mid x, y^i_{<j})\,/\,\pi_{\theta_\text{old}}(y^i_j \mid x, y^i_{<j})$ is the importance ratio for token $j$. GRPO optionally adds a KL penalty to the reference model, weighted by $\beta$.

\paragraph{Self-Distillation Fine-Tuning (SDFT).}
\looseness=-10000
For on-policy distillation, we use SDFT~\citep{sdft}, in which the model is its own teacher. The teacher's weights $\theta_T$ are an exponential moving average of the student's weights $\theta$,
\begin{equation}
\theta_T \;\leftarrow\; (1-\gamma)\,\theta_T + \gamma\,\theta,
\label{eq:sdft-ema}
\end{equation}
\looseness=-10000
with $\gamma < 1$, and the teacher's prompt also includes $d$, a reference response to the same question, which the student never sees. For a response $y \sim \pi_\theta(\cdot \mid x)$ sampled by the student, the loss is the forward KL between the teacher's and the student's next-token distributions at each position:
\begin{equation}
\mathcal{L}_\text{SDFT}(\theta) \;=\; \mathbb{E}_{x,\,y \sim \pi_\theta(\cdot|x)}\sum_{j=1}^{|y|}\mathrm{KL}\!\Bigl(\pi_{\theta_T}(\cdot \mid d, x, y_{<j})\;\big\|\;\pi_\theta(\cdot \mid x, y_{<j})\Bigr).
\label{eq:sdft}
\end{equation}

\section{\ourmethod{}}
\label{sec:method}

\looseness=-10000
\ourmethodit{} parameterizes fine-tuning as one trainable scalar logit per pretrained weight at a chosen subset of linear layers, with the base model frozen. During training, a sigmoid of each logit is annealed toward a step function, and at deployment each logit becomes a single bit deciding whether the corresponding weight is kept or deleted (set to zero). In pruning terms, \acronymit{} performs learned unstructured pruning to fine-tune the model.

\paragraph{Notation.}
\looseness=-10000
Let $W \in \mathbb{R}^{d_\text{out} \times d_\text{in}}$ denote a target weight matrix in the pretrained model and $G \in \mathbb{R}^{d_\text{out} \times d_\text{in}}$ a trainable logit matrix of the same shape, where the sign of $G_{ij}$ at deployment decides whether $W_{ij}$ is kept (positive) or zeroed (negative). We mask one such $W$ per target layer, each with its own independent $G$. In all our experiments, the target is the gate projection of transformer blocks (a single MLP weight per block). We follow GLU-specific pruning methods by treating the gate/up/down pathway as the main coupled sparsification structure~\citep{llmpruner,dass}: in the Llama and Qwen2.5 backbones we evaluate, SwiGLU MLPs take the form $\mathrm{down}(\mathrm{SiLU}(\mathrm{gate}(x)) \odot \mathrm{up}(x))$, so the gate projection multiplicatively controls the intermediate channels routed through the down projection.

\paragraph{Forward pass.}
\looseness=-10000
At training step $t \in \{0, 1, \dots, T\}$ (with $T$ the total step count), the layer applies a sigmoid relaxation of the binary masking decision per weight:
\begin{equation}
\widetilde{W}(t) \;=\; W \,\odot\, \sigma\!\bigl(\alpha(t) \,G\bigr),\qquad \sigma(x) = \frac{1}{1+e^{-x}}.
\label{eq:train-fwd}
\end{equation}
\looseness=-10000
We call $\sigma(\alpha(t)\,G)$ the \emph{soft mask} and $\mathbf{1}[G > 0]$ the \emph{hard mask}. The sharpness $\alpha(t) > 0$ sets how steep the sigmoid is around $G = 0$. A smaller $\alpha$ widens the band of logits whose soft mask lies strictly between $0$ and $1$, and raising $\alpha$ pulls the soft mask onto the hard one by the end. We use a linear ramp from $\alpha_\text{start} = 10^3$ to $\alpha_\text{end} = 10^4$, which we call \emph{mask sharpening}:
\begin{equation}
\alpha(t) \;=\; \alpha_\text{start} \,+\, \tfrac{t}{T} (\alpha_\text{end} - \alpha_\text{start}).
\label{eq:alpha-schedule}
\end{equation}
\looseness=-10000
At evaluation, the layer applies the hard mask:
\begin{equation}
\widehat{W} \;=\; W \,\odot\, \mathbf{1}\!\bigl[G > 0\bigr].
\label{eq:eval-fwd}
\end{equation}
\looseness=-10000
Gradients are computed through the deterministic soft mask in Eq.~\eqref{eq:train-fwd}, while the hard mask is used for evaluation and for generating rollouts in on-policy runs. By the end of training, the two agree: deploying the hard mask instead of the soft one changes accuracy by at most $0.08$ points.

\paragraph{Wanda prior initialization.}
\looseness=-10000
We initialize $G$ from the Wanda one-shot pruning score~\citep{wanda}, an importance signal validated as a strong predictor of which base weights are safe to zero. The Wanda score of a weight is its magnitude times the norm of its input activation, $s_{ij} = |W_{ij}| \cdot \lVert X_j \rVert_2$, where $X_j$ collects the activations of input channel $j$ in one forward pass over calibration prompts from the task's training distribution. Specifically, we sample each gate logit from one of two Gaussians depending on whether the weight is above or below the $\eta$-quantile of $s$:
\begin{equation}
G_{ij} \;\sim\;
\begin{cases}
\mathcal{N}(0.03,\ 0.01^2) & \parbox[t]{0.5\linewidth}{\raggedright if $s_{ij}$ is in the top $1{-}\eta$ fraction (high importance),} \\[2pt]
\mathcal{N}(-0.03,\ 0.01^2) & \text{otherwise (low importance).}
\end{cases}
\label{eq:wanda-init}
\end{equation}
\looseness=-10000
At our default $\eta = 1/2048$, used in every main experiment, this zeros approximately $\eta\,\Phi(3) + (1-\eta)\,\Phi(-3) \approx 0.18\%$ of gate-projection weights under the hard mask at initialization, where $\Phi$ is the standard normal CDF.

\paragraph{Training objective.}
\looseness=-10000
We instantiate the policy $\pi_\theta$ of Section~\ref{sec:background} by replacing $W$ with $\widetilde{W}(t)$ and keeping pretrained weights elsewhere, so the gate logits $G$ are the only trainable parameters in the model. The training loss $\mathcal{L}_\text{task}(\widetilde{W}(t))$ is the regime's own objective from Eqs.~\eqref{eq:sft} to~\eqref{eq:sdft}, plus the MaskLLM sparse weight regularizer $-\lambda\,\lVert\widetilde{W}(t)\rVert_2^{2}$, which mildly biases the mask toward retaining higher-magnitude weights. We do not constrain $G$ to any fixed sparsity level, leaving the training process to decide it.
\begin{itemize}[leftmargin=1.2em,itemsep=0pt,topsep=2pt]
\item \textbf{SFT.} The loss is $\mathcal{L}_\text{SFT}$, with $\lambda = 10^{-5}$.
\item \textbf{GRPO.} The loss is $\mathcal{L}_\text{GRPO}$, with $\lambda = 10^{-5}$. Policy rollouts use the hard mask of Eq.~\eqref{eq:eval-fwd}, while gradients flow into $G$ through the soft mask of Eq.~\eqref{eq:train-fwd}.
\item \textbf{SDFT.} The loss is $\mathcal{L}_\text{SDFT}$, with $\lambda = 0$, since the EMA teacher already provides an anchor. Rollouts use the hard mask as in GRPO, and the EMA teacher of Eq.~\eqref{eq:sdft-ema} specializes to a teacher gate $G_T \leftarrow (1-\gamma)\,G_T + \gamma\,G$, so the teacher's frozen base weights $W$ stay the same as the student's.
\end{itemize}

\paragraph{Compact storage and ease of deployment.}
\looseness=-10000
The trained adapter is the hard mask $\mathbf{1}[G > 0]$ at each masked layer, which takes one bit per masked pretrained weight before compression. We bit-pack the booleans with NumPy and compress the bit stream with zstd~\citep{zstd}. Because fewer than $0.05\%$ of all weights are deleted in every GRPO run of Table~\ref{tab:perf-combined}, the bit stream is highly redundant, and zstd compresses it to $1.0$~MB at Llama1B and $8.4$~MB at Llama8B. This gives \acronymit{} a practical deployment advantage in compact storage and cheap adapter switching: changing tasks only loads a compact bitmask and scatters zeros into the base weights, rather than loading a separate full model checkpoint.

\paragraph{Connection to MaskLLM.}
\looseness=-10000
MaskLLM~\citep{maskllm} learns $2{:}4$ masks for inference speedup with a Gumbel-Softmax selector over $N{:}M$ candidate patterns. \acronymit{} has a different goal, expanding model capability, and a different form, with one deterministic logit per weight and no fixed sparsity pattern (Appendix~\ref{app:maskllm-reduction} relates the two formally). \acronymit{} also adopts MaskLLM's Wanda prior initialization (Eq.~\eqref{eq:wanda-init}) and its sparse weight regularizer.

\section{Experiments}
\label{sec:experiment}

\subsection{Experimental Setup}
\label{sec:experimentsetup}

\paragraph{Tasks, benchmarks, and models.}
\looseness=-10000
We study three post-training regimes on two model families, referring to a family as Llama or Qwen and to a size by its suffix (e.g., Llama1B, Qwen7B). Evaluation uses greedy decoding, and we report the best validation accuracy across saved checkpoints.
\begin{itemize}
\item \looseness=-10000
\textbf{GRPO, in distribution.} We train and evaluate on GSM8K~\citep{gsm8k} with Llama1B, Llama3B, and Llama8B (Llama3.2-1B/3B-Instruct and Llama3-8B-Instruct)~\citep{llama3}, validating every $25$ optimizer steps.
\item \looseness=-10000
\textbf{GRPO, out of distribution.} Following SimpleRL-Zoo~\citep{simplerlzoo}, we train on its math corpus and report the mean over six held-out benchmarks (GSM8K, MATH500, Minerva Math, OlympiadBench, AIME24, AMC23; per-benchmark results in Appendix~\ref{app:qwen-perbench}) with Qwen0.5B, Qwen3B, and Qwen7B (Qwen2.5 base)~\citep{qwen25}. At larger scale, we add Qwen14B (Qwen2.5-14B) and the Qwen3-30B-A3B mixture-of-experts model ($30$B total and $3$B active parameters). We validate every $10$ optimizer steps.
\item \looseness=-10000
\textbf{SFT.} We train Llama1B and Llama3B (Llama3.2-1B/3B-Instruct) separately on each of the $14$ MMLU-Pro~\citep{mmlupro} subjects and $23$ BBEH~\citep{bbeh} subtasks, validate at the end of training, and average accuracy within each suite.
\item \looseness=-10000
\textbf{SDFT.} We follow the ToolAlpaca~\citep{toolalpaca} tool-use and Science settings of~\citet{sdft} with Qwen7B (Qwen2.5-7B-Instruct), reporting exact-match accuracy under each task's evaluator. Science is the Chemistry L-3 subset of SciKnowEval~\citep{sciknoweval}. We validate every $100$ optimizer steps on ToolAlpaca and every $25$ on Science.
\end{itemize}

\paragraph{Training pipelines.}
\looseness=-10000
All runs use AdamW with fp32 master weights, bf16 mixed precision, and gradient checkpointing.
\begin{itemize}
\item \looseness=-10000
\textbf{GRPO.} We use verl~\citep{verl} with a constant learning rate. GSM8K uses prompt batch $64$, $n{=}4$ rollouts, response length $4096$, $3$ epochs, and no KL penalty, while SimpleRL-Zoo uses prompt batch $256$, $n{=}8$ rollouts, response length $4096$, $150$ steps, and KL penalty $10^{-4}$.
\item \looseness=-10000
\textbf{SFT.} We minimize cross-entropy with a constant learning rate for $T{=}400$ steps per task at effective batch $4$, with a $75/25$ train/eval split (per-task counts in Appendix~\ref{app:sft-splits}).
\item \looseness=-10000
\textbf{SDFT.} We follow~\citet{sdft}: cosine schedule with $0.1$ warmup ratio, prompt batch $32$, $2$ epochs, EMA mixup $\gamma = 0.01$, and one demonstration prepended to the teacher's input.
\end{itemize}

\paragraph{Baselines.}
\looseness=-10000
All three baselines use the same data and number of steps as \acronymit{}.
\begin{itemize}
\item \looseness=-10000
\textbf{FFT.} Full fine-tuning trains all weights and is the reference we aim to match. Its learning rate is $10^{-6}$ on GSM8K, $5{\times}10^{-7}$ on SimpleRL-Zoo, $10^{-6}$ to $3{\times}10^{-6}$ on SFT (Appendix~\ref{app:sft-fft-sweep}), and $5{\times}10^{-5}$ on SDFT as in~\citet{sdft}.
\item \looseness=-10000
\textbf{LoRA.} We train rank-$1$ and rank-$4$ adapters~\citep{lora} on all seven projections in each block (q/k/v/o and gate/up/down) with learning rate $10^{-4}$.
\item \looseness=-10000
\textbf{MTH.} The closest sparse-update method~\citep{mth} trains a random $1\%$ or $0.05\%$ of all parameters ($99\%$ and $99.95\%$ sparsity), with the learning rates recommended by~\citet{mth} ($10^{-4}$ and $10^{-3}$).
\end{itemize}

\looseness=-10000
\paragraph{\acronym{} implementation.}
\looseness=-10000
We compute the Wanda scores once per task, on $2048$ training prompts for GRPO and SFT and on $32$ for SDFT.
\begin{itemize}
\item \looseness=-10000
\textbf{GRPO.} Every main run uses $\eta = 1/2048$, learning rate $3{\times}10^{-4}$, $(\alpha_\text{start}, \alpha_\text{end}) = (10^3, 10^4)$, and $\lambda = 10^{-5}$, with no tuning per benchmark or model size (Section~\ref{sec:hp-sensitivity} varies $\eta$, learning rate, and $\lambda$).
\item \looseness=-10000
\textbf{SFT.} We keep the same $\eta$ and $\lambda$, with learning rate $5{\times}10^{-4}$ and $(\alpha_\text{start}, \alpha_\text{end}) = (10^2, 10^3)$ for every subject and subtask.
\item \looseness=-10000
\textbf{SDFT.} We set $\eta = 1/2048$ and $\lambda = 0$ and use a hard teacher mask, with learning rate $5{\times}10^{-3}$ and $(10^2, 10^4)$ on ToolAlpaca and $3{\times}10^{-3}$ and $(10^2, 10^3)$ on Science.
\end{itemize}

\paragraph{Hardware.}
\looseness=-10000
We train on H200 GPUs, using one GPU for backbones up to $3$B and multiple GPUs with FSDP~\citep{fsdp} for larger models. SFT and the post-hoc analyses use L40S GPUs.

\subsection{Results on GRPO Post-Training}
\label{sec:results}

\looseness=-10000
We train \acronymit{} and three baselines (FFT, LoRA at ranks $1$ and $4$, MTH at $99\%$ and $99.95\%$) on the two GRPO settings of Section~\ref{sec:experimentsetup}. Table~\ref{tab:perf-combined} reports best-checkpoint validation accuracy and bytes-on-disk adapter size for six backbones from $0.5$B to $8$B across two model families. Every \acronymit{} run deletes fewer than $0.05\%$ of all weights (Appendix~\ref{app:per-cell-sparsity}). Appendix~\ref{app:size-accounting} defines adapter size per method.
\begin{table}[t]
\caption{
\looseness=-10000
\textbf{\acronymit{} has the smallest adapter at every scale and matches FFT at the largest backbone of each family.} Best-checkpoint validation accuracy (\%) and adapter size (MB) under GRPO, on GSM8K for Llama and on the six-benchmark SimpleRL-Zoo mean for Qwen. The baselines are full fine-tuning (FFT), low-rank adapters (LoRA), and random sparse fine-tuning (MTH).
}
\label{tab:perf-combined}
\centering
\setlength{\tabcolsep}{2.5pt}
\small
\begin{tabular}{l rr >{\hspace{3pt}}rr >{\hspace{3pt}}rr >{\hspace{3pt}}rr >{\hspace{3pt}}rr >{\hspace{3pt}}rr}
\toprule
                & \multicolumn{6}{c}{\textbf{Llama, GSM8K}}                                                                                & \multicolumn{6}{c}{\textbf{Qwen, SimpleRL-Zoo}} \\
\cmidrule(lr){2-7}\cmidrule(lr){8-13}
                & \multicolumn{2}{c}{\textbf{1B}} & \multicolumn{2}{c}{\textbf{3B}} & \multicolumn{2}{c}{\textbf{8B}} & \multicolumn{2}{c}{\textbf{0.5B}} & \multicolumn{2}{c}{\textbf{3B}} & \multicolumn{2}{c}{\textbf{7B}} \\
\cmidrule(lr){2-3}\cmidrule(lr){4-5}\cmidrule(lr){6-7}\cmidrule(lr){8-9}\cmidrule(lr){10-11}\cmidrule(lr){12-13}
\textbf{Method} & {\footnotesize\textbf{Acc.}} & {\footnotesize\textbf{Size}} & {\footnotesize\textbf{Acc.}} & {\footnotesize\textbf{Size}} & {\footnotesize\textbf{Acc.}} & {\footnotesize\textbf{Size}} & {\footnotesize\textbf{Acc.}} & {\footnotesize\textbf{Size}} & {\footnotesize\textbf{Acc.}} & {\footnotesize\textbf{Size}} & {\footnotesize\textbf{Acc.}} & {\footnotesize\textbf{Size}} \\
\midrule
Base            & 1.21           & 0 & 53.53          & 0 & 60.27          & 0 & 10.44          & 0 & 37.73          & 0 & 40.18          & 0 \\
FFT             & 60.50          & 4714 & \textbf{82.49} & 12256 & 83.24          & 30633 & 19.01          & 1885 & \textbf{44.35} & 11772 & 51.89          & 29051 \\
LoRA r=1        & 59.29          & 2.7 & 82.39          & 5.8 & 83.38          & 10.0 & 20.30          & 2.1 & 43.55          & 7.1 & 52.13          & 9.6 \\
LoRA r=4        & 60.58          & 10.8 & 82.44          & 23.2 & 82.87          & 40.0 & 20.30          & 8.4 & 43.59          & 28.6 & 51.70          & 38.5 \\
MTH 99\%        & \textbf{60.73} & 68.2 & 80.53          & 177.4 & 83.31          & 443.6 & \textbf{21.84} & 27.3 & 44.16          & 170.4 & 52.03          & 420.7 \\
MTH 99.95\%     & 58.61          & 4.1 & 82.17          & 10.7 & 83.07          & 26.6 & 20.16          & 1.7 & 43.87          & 13.7 & 51.24          & 25.3 \\
\rowcolor{TableHighlight}
\acronymit{}    & 54.44          & \textbf{1.0} & 81.49          & \textbf{3.1} & \textbf{83.55} & \textbf{8.4} & 17.64          & \textbf{0.5} & 41.98          & \textbf{2.7} & \textbf{52.32} & \textbf{6.9} \\
\bottomrule
\end{tabular}
\end{table}

\looseness=-10000
\begin{table}[t]
\caption{
\looseness=-10000
\textbf{\acronymit{} matches FFT on dense and mixture-of-experts models from 7B to 30B with an adapter thousands of times smaller.} Best-checkpoint validation accuracy (\%) and adapter size (MB) under GRPO. Qwen14B and Qwen3-30B-A3B use the SimpleRL-Zoo setup of Qwen7B.
}
\label{tab:largest}
\centering
\setlength{\tabcolsep}{4pt}
\small
\begin{tabular}{l rr >{\hspace{4pt}}rr >{\hspace{4pt}}rr >{\hspace{4pt}}rr}
\toprule
& \multicolumn{2}{c}{\textbf{Llama8B}} & \multicolumn{2}{c}{\textbf{Qwen7B}} & \multicolumn{2}{c}{\textbf{Qwen14B}} & \multicolumn{2}{c}{\textbf{Qwen3-30B-A3B}} \\
& \multicolumn{2}{c}{\footnotesize GSM8K} & \multicolumn{2}{c}{\footnotesize SimpleRL-Zoo} & \multicolumn{2}{c}{\footnotesize SimpleRL-Zoo} & \multicolumn{2}{c}{\footnotesize SimpleRL-Zoo} \\
\cmidrule(lr){2-3}\cmidrule(lr){4-5}\cmidrule(lr){6-7}\cmidrule(lr){8-9}
\textbf{Method} & {\footnotesize\textbf{Acc.}} & {\footnotesize\textbf{Size}} & {\footnotesize\textbf{Acc.}} & {\footnotesize\textbf{Size}} & {\footnotesize\textbf{Acc.}} & {\footnotesize\textbf{Size}} & {\footnotesize\textbf{Acc.}} & {\footnotesize\textbf{Size}} \\
\midrule
Base & 60.27 & 0 & 40.18 & 0 & 44.37 & 0 & 49.41 & 0 \\
FFT & 83.24 & 30633 & 51.89 & 29051 & 54.87 & 56343 & 58.43 & 116471 \\
\rowcolor{TableHighlight}
\acronymit{} & \textbf{83.55} & \textbf{8.4} & \textbf{52.32} & \textbf{6.9} & \textbf{55.43} & \textbf{11.4} & \textbf{58.54} & \textbf{29.6} \\
\bottomrule
\end{tabular}
\end{table}

\paragraph{\acronymit{} matches FFT at 7B and 8B.}
\looseness=-10000
At the largest backbone in each family, \acronymit{} scores $83.55\%$ against FFT's $83.24\%$ on Llama8B GSM8K and $52.32\%$ against $51.89\%$ on Qwen7B SimpleRL-Zoo. Tuning FFT's learning rate and KL coefficient does not change this (Appendix~\ref{app:fft-sweep}). At these two backbones, \acronymit{} is also competitive with both LoRA ranks and both MTH sparsity levels (Table~\ref{tab:perf-combined}). Since \acronymit{} can only delete pretrained weights, this restriction may act as a mild regularizer, which could explain why \acronymit{} matches FFT here. The same restriction pays off more clearly for retention, where \acronymit{} keeps the prior-task accuracy that FFT loses (Section~\ref{sec:retention}).

\paragraph{The gap to FFT closes with scale.}
\looseness=-10000
At smaller backbones, \acronymit{} recovers $90\%$ to $99\%$ of FFT's accuracy. At larger scale, \acronymit{} matches FFT on Qwen14B and on the Qwen3-30B-A3B mixture-of-experts model (Table~\ref{tab:largest}). For the mixture-of-experts model, we mask the gate projection inside each of its $128$ experts in every layer and leave the router unchanged. These results suggest that deletion becomes more effective as models grow.

\begin{table}[t]
\caption{
\looseness=-10000
\textbf{\acronymit{} acquires new capabilities.} Pass@k (\%), the share of problems solved within $k$ attempts, estimated from $128$ samples per problem at temperature $1.0$. A lead at large $k$ means solving problems the base model misses even with many attempts.
}
\label{tab:passk}
\centering
\setlength{\tabcolsep}{4pt}
\small
\begin{tabular}{l rrrrr rrrrr}
\toprule
& \multicolumn{5}{c}{\textbf{Llama8B, GSM8K}} & \multicolumn{5}{c}{\textbf{Qwen7B, SimpleRL-Zoo}} \\
\cmidrule(lr){2-6}\cmidrule(lr){7-11}
\textbf{Method} & {\footnotesize$k{=}1$} & {\footnotesize$2$} & {\footnotesize$8$} & {\footnotesize$32$} & {\footnotesize$128$} & {\footnotesize$k{=}1$} & {\footnotesize$2$} & {\footnotesize$8$} & {\footnotesize$32$} & {\footnotesize$128$} \\
\midrule
Base & 53.20 & 66.75 & 83.64 & 92.40 & 95.29 & 27.35 & 38.75 & 56.18 & 66.26 & 74.01 \\
\rowcolor{TableHighlight}
\acronymit{} & \textbf{81.75} & \textbf{88.51} & \textbf{95.75} & \textbf{98.44} & \textbf{99.62} & \textbf{49.39} & \textbf{53.22} & \textbf{64.64} & \textbf{73.23} & \textbf{80.08} \\
\bottomrule
\end{tabular}
\end{table}

\paragraph{\acronymit{} acquires new capabilities.}
\looseness=-10000
Prior sparsity methods aim to preserve base performance under aggressive pruning~\citep{maskllm, wanda, sparsegpt}. \acronymit{} instead uses sparsity as the learning mechanism itself. On pass@k (Table~\ref{tab:passk}), \acronymit{} leads the base model at every $k$. Even at $k{=}128$, the base model's pass@k stays $4.3$ and $6.1$ points below \acronymit{}'s on Llama8B and Qwen7B.

\paragraph{Smallest adapter at every scale.}
\looseness=-10000
\acronymit{}'s adapter is $0.5$~MB at Qwen0.5B and $8.4$~MB at Llama8B, $3{,}600$ to $4{,}700{\times}$ smaller than the FFT checkpoints and the smallest of any adapter family at every scale. The closest baselines by size are LoRA $r{=}1$ and MTH at $99.95\%$ sparsity. They are still often several times larger than \acronymit{}, and MTH at $99\%$ is more than $50{\times}$ larger in every run. After compression, the mask takes about $0.03$ bits per target weight, while every other method stores floating-point values for the weights it changes.

\subsection{Learning the Mask Drives the Gains}
\label{sec:wanda-vs-base}

\looseness=-10000
Do \acronymit{}'s gains over the base model come from learning the mask itself, beyond the Wanda prior that initializes the gates? We test this directly. For each run in Table~\ref{tab:perf-combined}, we apply Wanda one-shot pruning without GRPO training at $8$ sparsity levels from $1/2$ to $1/8192$. We evaluate each level on the same validation set and record the \emph{best} accuracy (Appendix~\ref{app:wanda-sweep} reports every level). Table~\ref{tab:wanda-vs-base} compares this best-Wanda score with the base model and the trained \acronymit{} mask.

\begin{table}[t]
  \centering
  \caption{
\looseness=-10000
  \textbf{Wanda alone stays near the base model.} Validation accuracy (\%) of the base model, of Wanda one-shot pruning at the best of eight sparsity levels, and of the best checkpoint of the \acronymit{} mask trained with GRPO.
  }
  \label{tab:wanda-vs-base}
  \setlength{\tabcolsep}{4pt}
  \small
  \begin{tabular}{l rrr rrr}
  \toprule
  & \multicolumn{3}{c}{{\textbf{Llama, GSM8K}}}
  & \multicolumn{3}{c}{{\textbf{Qwen, SimpleRL-Zoo}}} \\
  \cmidrule(lr){2-4}\cmidrule(lr){5-7}
  \textbf{Method}     & 1B    & 3B     & 8B    & 0.5B  & 3B    & 7B    \\
  \midrule
  Base                & 1.21  & 53.53  & 60.27 & 10.44 & 37.73 & 40.18 \\
  Wanda (best $\eta$) & 1.59  & 52.46  & 60.65 & 11.42 & 38.70 & 42.49 \\
  \rowcolor{TableHighlight}
  \acronymit{}        & \textbf{54.44} & \textbf{81.49} & \textbf{83.55} & \textbf{17.64} & \textbf{41.98} & \textbf{52.32} \\
  \bottomrule
  \end{tabular}
\end{table}

\looseness=-10000
The best Wanda mask stays within $2.4$ points of the base on all six backbones, while \acronymit{}, starting from the same Wanda scores, gains up to $53.2$ points.

\subsection{\acronym{} Forgets Less}
\label{sec:retention}

\begin{figure}[t]
  \centering
  \includegraphics[width=\linewidth]{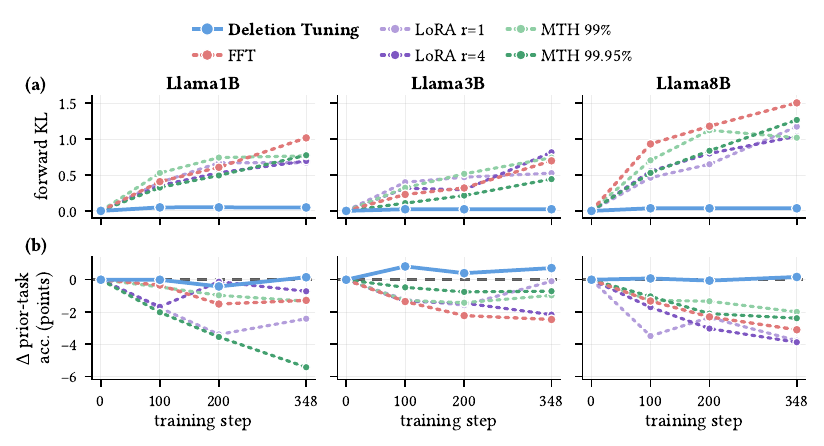}
  \caption{
\looseness=-10000
  \textbf{\acronymit{} keeps prior-task accuracy at every Llama scale while the baselines drift.} (a) Forward KL between the trained and base policies on GSM8K during GRPO training, the forgetting predictor of~\citet{rlrazor}. (b) Change in mean accuracy on a six-task prior-task suite from the base model (dashed line).
  }
  \label{fig:retention}
\end{figure}

\looseness=-10000
Forward KL on the new task predicts how much a fine-tuned model forgets~\citep{rlrazor}. We hypothesize that deletion keeps the perturbation to the base policy small, because each edit can only set a weight to zero and training chooses which few weights to edit, yielding low forward KL and therefore little forgetting. Following~\citet{rlrazor}'s evaluation, Figure~\ref{fig:retention} tracks forward KL and prior-task accuracy over training for FFT, LoRA, MTH, and \acronymit{} at the three Llama scales. The prior-task suite has six benchmarks: HellaSwag, MMLU, TruthfulQA, WinoGrande, IFEval, and HumanEval.

\looseness=-10000
By step $348$, FFT's forward KL reaches $1.0$ on Llama1B and $1.5$ on Llama8B, and its prior-task accuracy ends $1.3$ to $3.1$ points below base. \acronymit{}'s forward KL stays at or below $0.05$ at all three scales, and its prior-task accuracy stays within one point of base throughout training. LoRA and MTH train only a small fraction of parameters, yet their forward KL reaches $0.4$ to $1.3$ and their prior-task accuracy falls up to $5.4$ points below base. MTH at $99.95\%$ sparsity trains $0.05\%$ of all parameters, comparable to the $0.035\%$ to $0.046\%$ that \acronymit{} deletes at these scales (Table~\ref{tab:per-cell-sparsity}), suggesting that the small number of edited weights does not explain \acronymit{}'s retention, and that deletion itself keeps forgetting low.

\subsection{Why \acronym{} Works}

\label{sec:geometry}

\looseness=-10000
How can binary masking alone (Section~\ref{sec:results}) match FFT under GRPO at scale? We compare the weight updates of \acronymit{} and the three baselines at a saved checkpoint on three geometric measures.

\paragraph{Update geometry.}
\looseness=-10000
Let $W_0$ denote the pretrained value of a weight matrix, $W_+$ its effective deployed value at the saved checkpoint (Llama8B step $348$, Qwen7B step $100$), and $\Delta W = W_+ - W_0$ the weight delta. For FFT, $W_+$ is the trained matrix, and for the adapters it is the merged or masked matrix that runs at inference. We report three quantities, each averaged over all attention and MLP linear layers:
\begin{itemize}
\item \looseness=-10000
\textbf{Stable rank} of $\Delta W$ is $\lVert \Delta W \rVert_F^2 / \lVert \Delta W \rVert_2^2 = \sum_{m=1}^{q}\sigma_m^2 / \sigma_1^2$, where $q=\min(d_{\text{out}}, d_{\text{in}})$ and $\sigma_1 \geq \cdots \geq \sigma_q$ are the singular values of $\Delta W$. It counts how many directions the update uses, independent of its scale: it is close to $1$ when one direction dominates and close to $k$ when $k$ directions carry similar energy.
\item \looseness=-10000
\textbf{Spectrum drift} is the normalized singular-value shift $\lVert \sigma(W_+) - \sigma(W_0) \rVert_2 / \lVert \sigma(W_0) \rVert_2$~\citep{pathnottaken}, where $\sigma(W)$ lists the singular values of $W$. It is small when training leaves the singular values close to where they started.
\item \looseness=-10000
\textbf{Principal-weight overlap} measures the fraction of changed weight positions that fall on what~\citet{pathnottaken} call the base model's \emph{principal weights}. The \emph{principal-weight mask}
\[ M_{\text{princ}} = \text{Top}_{50\%}\bigl(\lvert W_0^{(128)} \rvert\bigr) \]
marks the top $50\%$ entries by magnitude of $W_0^{(128)}$, the best rank-$128$ approximation of $W_0$. The \emph{update mask} $M_{\text{update}}$ marks positions where the deployed matrix differs from the base by
\[ \lvert W_+(i,j) - W_0(i,j) \rvert > 10^{-3} \cdot \max(\lvert W_0(i,j) \rvert, \lvert W_+(i,j) \rvert), \]
the relative tolerance of~\citet{pathnottaken}, which ignores changes too small to register in bf16. The overlap is $\lvert M_{\text{princ}} \cap M_{\text{update}}\rvert / \lvert M_{\text{update}}\rvert$. Update positions spread uniformly at random would score $0.5$, the density of $M_{\text{princ}}$.
\end{itemize}

\begin{figure}[t]
\centering
\includegraphics[width=\linewidth]{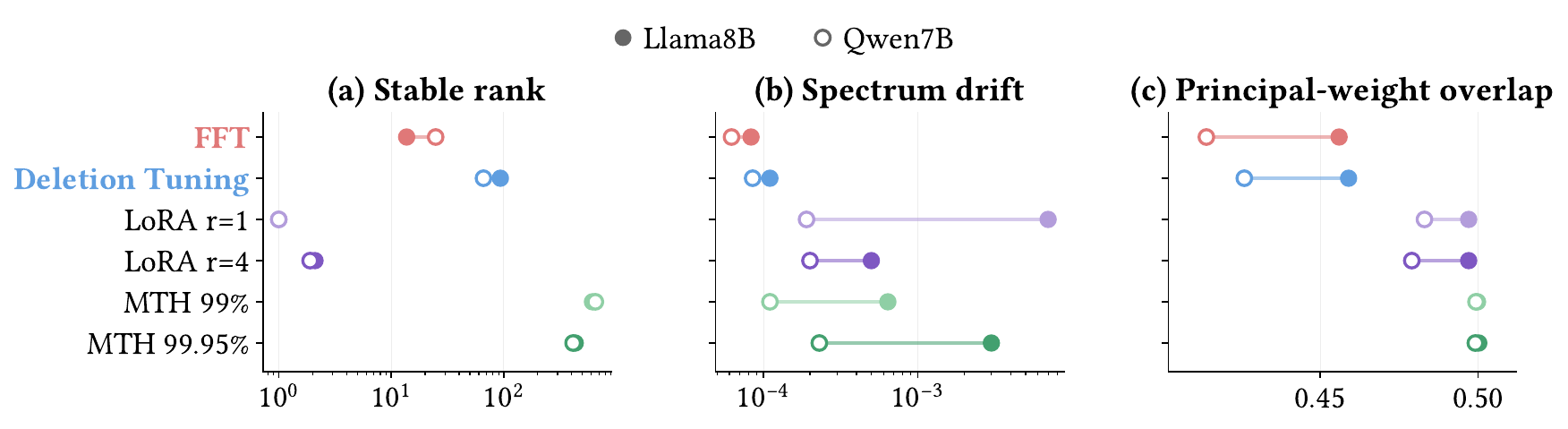}
\caption{
\looseness=-10000
\textbf{\acronymit{} (blue) tracks FFT (red) most closely across three update-geometry metrics.} Values at the saved GRPO checkpoint for Llama8B (filled) and Qwen7B (open): (a) stable rank of $\Delta W$, (b) spectrum drift, and (c) principal-weight overlap. Overlap below the random baseline of $0.5$ is the RLVR signature reported by~\citet{pathnottaken}.
}
\label{fig:geometry}
\end{figure}

\looseness=-10000
Figure~\ref{fig:geometry} shows the three metrics for \acronymit{} and every baseline setting, and \acronymit{} is the closest to FFT on the three together. Its stable rank of $66$ to $93$ sits far below MTH's $410$ to $650$ and above LoRA's $1$ to $2$. Its spectrum drift stays within $1.4{\times}$ of FFT's, while LoRA $r{=}1$ drifts up to $84{\times}$ more than FFT on Llama8B. Its principal-weight overlap of $0.43$ to $0.46$ matches FFT's sub-random $0.41$ to $0.46$. LoRA stays near the random baseline at $0.48$ to $0.50$, and MTH sits at $0.50$.

\paragraph{Interpretation.}
\looseness=-10000
Low spectrum drift and sub-random principal-weight overlap are the RLVR signature reported by~\citet{pathnottaken}: RLVR leaves the singular values of each weight matrix nearly unchanged and steers its updates away from the principal weights. LoRA caps the rank of its update at $r$, and MTH trains weights chosen at random before training. \acronymit{} has neither restriction, since it learns which weights to change and has no rank cap. These two freedoms help explain how \acronymit{} reaches FFT-like geometry even though it never sets a weight to a new non-zero value.

\subsection{Generalizing Beyond GRPO: Results on SFT and SDFT}
\label{sec:results-sft-sdft}

\looseness=-10000
Does deletion also work when the training signal is denser or comes from outside the model? GRPO samples from the model itself and rewards whole responses, and RL updates of this kind are known to be sparse~\citep{rlsmallsubnet}. SFT instead imitates fixed external targets token by token and produces dense updates~\citep{rlsmallsubnet}, which makes it a harder test for an adapter that can only delete. SDFT sits in between: the student samples its own responses, and a teacher supplies a target distribution at every token. We test \acronymit{} with SFT on knowledge and reasoning benchmarks (MMLU-Pro and BBEH) and with SDFT on tool use (ToolAlpaca) and science questions (Science), using the setups of Section~\ref{sec:experimentsetup}.

\looseness=-10000
Under SFT, \acronymit{} matches FFT at 3B on both MMLU-Pro and BBEH (Table~\ref{tab:perf-sft}). Under SDFT, \acronymit{} matches FFT on both Qwen7B tasks (Table~\ref{tab:perf-sdft}). In both regimes, its adapters are $600$ to $4{,}700{\times}$ smaller than FFT checkpoints.

\begin{table}[t]
\caption{
\looseness=-10000
\textbf{Under SFT, \acronymit{} matches FFT at 3B.} Best-checkpoint evaluation accuracy (\%), averaged over the $14$ MMLU-Pro subjects and the $23$ BBEH subtasks, and adapter size (MB) for Llama1B and Llama3B.
}
\label{tab:perf-sft}
\centering
\setlength{\tabcolsep}{4pt}
\small
\begin{tabular}{l rr >{\hspace{4pt}}rr >{\hspace{4pt}}rr >{\hspace{4pt}}rr}
\toprule
                & \multicolumn{4}{c}{\textbf{MMLU-Pro}}                            & \multicolumn{4}{c}{\textbf{BBEH}} \\
\cmidrule(lr){2-5}\cmidrule(lr){6-9}
                & \multicolumn{2}{c}{\textbf{1B}} & \multicolumn{2}{c}{\textbf{3B}} & \multicolumn{2}{c}{\textbf{1B}} & \multicolumn{2}{c}{\textbf{3B}} \\
\cmidrule(lr){2-3}\cmidrule(lr){4-5}\cmidrule(lr){6-7}\cmidrule(lr){8-9}
\textbf{Method} & {\footnotesize\textbf{Acc.}} & {\footnotesize\textbf{Size}} & {\footnotesize\textbf{Acc.}} & {\footnotesize\textbf{Size}} & {\footnotesize\textbf{Acc.}} & {\footnotesize\textbf{Size}} & {\footnotesize\textbf{Acc.}} & {\footnotesize\textbf{Size}} \\
\midrule
Base            & 10.37          & 0 & 12.63          & 0 &  2.17         & 0 &  8.87           & 0 \\
FFT             & 16.97          & 4714 & 21.19          & 12256 & \textbf{8.35} & 4714 & \textbf{11.39}  & 12256 \\
\rowcolor{TableHighlight}
\acronymit{}    & \textbf{18.62} & \textbf{1.0} & \textbf{22.28} & \textbf{3.1} &  5.65         & \textbf{1.0} & \textbf{11.39}  & \textbf{3.1} \\
\bottomrule
\end{tabular}
\end{table}

\begin{table}[t]
\caption{
\looseness=-10000
\textbf{Under SDFT, \acronymit{} matches FFT on Qwen7B.} Best-checkpoint exact-match accuracy (\%) and adapter size (MB) on ToolAlpaca tool use and Science.
}
\label{tab:perf-sdft}
\centering
\setlength{\tabcolsep}{5pt}
\small
\begin{tabular}{l rr rr}
\toprule
                      & \multicolumn{2}{c}{\textbf{ToolAlpaca}} & \multicolumn{2}{c}{\textbf{Science}} \\
\cmidrule(lr){2-3}\cmidrule(lr){4-5}
\textbf{Method}       & {\footnotesize\textbf{Acc.}} & {\footnotesize\textbf{Size}} & {\footnotesize\textbf{Acc.}} & {\footnotesize\textbf{Size}} \\
\midrule
Base                  & 41.24          & 0 & 32.35          & 0 \\
FFT                   & 68.04          & 29051 & \textbf{71.20} & 29051 \\
\rowcolor{TableHighlight}
\acronymit{}          & \textbf{69.07} & \textbf{27.1} & 70.61          & \textbf{47.1} \\
\bottomrule
\end{tabular}
\vspace{-5pt}
\end{table}

\subsection{\acronym{} Is Robust to Hyperparameter Changes}
\label{sec:hp-sensitivity}

\looseness=-10000
\acronymit{} works across a range of values for its hyperparameters: initial sparsity $\eta$, learning rate $\mathit{lr}$, and sparse-weight regularization strength $\lambda$. We sweep $\eta \in \{1/2048, 1/4096, 1/8192\}$, $\mathit{lr} \in \{10^{-4}, 3{\times}10^{-4}\}$, and $\lambda \in \{0, 10^{-5}\}$ on Llama8B GSM8K and Qwen7B SimpleRL-Zoo (Tables~\ref{tab:hp-sweep-llama} and~\ref{tab:hp-sweep-qwen} in Appendix~\ref{app:hp-sweep}). Every setting lands within $1.6$ points of the best run on Llama8B and within $3.2$ points on Qwen7B, and the single configuration we use for all main runs is at or near the best setting at both backbones. Varying the initial sparsity $\eta$ by $4{\times}$ changes accuracy by at most $0.8$ points on Llama8B and $1.7$ points on Qwen7B at fixed $\mathit{lr}$ and $\lambda$, so the default configuration is a reasonable starting point for new tasks.

\vspace{-4pt}
\section{Conclusion}
\label{sec:conclusions}

\looseness=-10000
We introduced \ourmethodit{}, which fine-tunes a model by learning only a binary mask over its gate projections. At the largest scale we test in each regime (Qwen14B and the Qwen3-30B-A3B mixture-of-experts model under GRPO, Qwen7B under SDFT, and Llama3B under SFT), it matches full fine-tuning. The mask is smaller than a rank-$1$ LoRA and hundreds to thousands of times smaller than an FFT checkpoint. In our GRPO retention runs, \acronymit{} keeps prior-task accuracy near the base level throughout training, while FFT, LoRA, and MTH lose up to $5.4$ points. Its weight updates also resemble FFT's more closely than those of LoRA or MTH across stable rank, spectrum drift, and principal-weight overlap, which helps explain how deletion alone can match FFT. These results suggest that post-training can work by selection alone, keeping nearly every pretrained weight and choosing the few to delete.

\paragraph{Limitations and Future Work.}
\looseness=-10000
Because \acronymit{} forgets little, a natural next test is continual learning over a sequence of tasks. Training keeps one logit per masked weight, so \acronymit{} uses less training memory than FFT but more than LoRA. Restricting the logits to a small set of candidate weights could close this gap. The masks are also small and discrete, which aids interpretability: comparing which weights different tasks delete could localize the changes that post-training makes to a model. Masks for different tasks might also be combined, for example by taking their union, to merge skills without further training. Finally, restricting the mask to $N{:}M$, channel-wise, or block patterns could let one mask both adapt and compress the model.

\section*{Acknowledgments}
\looseness=-10000
We thank members of the NYU Agentic Learning AI Lab and the Calgary ML Lab for helpful discussions. JL is supported by an NSERC PGS-D Scholarship. This work was supported in part by Visko Platform, Pionex, a Google TPU Award, a Google ML and Systems Junior Faculty Award, and the Institute of Information \& Communications Technology Planning \& Evaluation (IITP) with a grant funded by the Ministry of Science and ICT (MSIT) of the Republic of Korea in connection with the Global AI Frontier Lab International Collaborative Research Agreement (No.~RS-2024-00469482 \& RS-2024-00509279). Computational support was provided by NYU High Performance Computing through its resources, services, and staff expertise. This research was enabled in part by support provided by the Digital Research Alliance of Canada (alliancecan.ca).

ML, TP, and YI gratefully acknowledge the support of Alberta Innovates (ALLRP 577350-22, ALLRP 600038-24), the Natural Sciences and Engineering Research Council of Canada (NSERC) (RGPIN-2022-03120, DGECR-2022-00358), Defence Research and Development Canada (DGDND-2022-03120), and NSERC/Agence Nationale de la Recherche (ANR) (ALLRP 602719-24). This project was undertaken thanks to funding from IVADO and the Canada First Research Excellence Fund. ML additionally acknowledges the support of RBC Borealis through the RBC Borealis AI Global Fellowship Award. 

\bibliographystyle{apalike}
\bibliography{ref}

\clearpage
\appendix

\section{Adapter Size}
\label{app:size-accounting}

\looseness=-10000
The \textbf{Size} entries in Tables~\ref{tab:perf-combined},~\ref{tab:largest},~\ref{tab:perf-sft}, and~\ref{tab:perf-sdft} (in binary units, 1\,GB${}=1024$\,MB) are lower bounds on the bytes on disk needed to rebuild the trained model from the public base checkpoint, computed the same way for every method. We leave out training overhead that deployment discards (file headers, the Adam moments $m$ and $v$, and learning-rate scheduler and RNG state), which would otherwise dominate the smallest adapters.

\looseness=-10000
\paragraph{Full fine-tuning (FFT).} A fine-tuned LLM checkpoint at fp32:
\looseness=-10000
\[
    \text{size}(\text{FFT}) = N_\text{total} \cdot 4\,\text{bytes},
\]
\looseness=-10000
where $N_\text{total}$ is the total parameter count. We use fp32 to match the trained master weights, which we store in fp32 for every adapter family.

\looseness=-10000
\paragraph{LoRA.} The $B$ and $A$ low-rank factor matrices, no bias terms:
\looseness=-10000
\[
    \text{size}(\text{LoRA-}r) = \sum_{\ell \in \mathcal{T}_\text{LoRA}} \bigl(r \cdot d_\text{in}^{(\ell)} + d_\text{out}^{(\ell)} \cdot r\bigr) \cdot 4\,\text{bytes}.
\]

\looseness=-10000
\paragraph{MTH.} The active fp32 values plus a compressed boolean mask of their positions:
\looseness=-10000
\[
    \text{size}(\text{MTH}) = N_\text{active} \cdot 4\,\text{bytes} + \text{zstd}\bigl(\text{packbits}(\mathbf{1}[\text{active}])\bigr),
\]
\looseness=-10000
where $N_\text{active} = (1-s) \cdot N_\text{total}$ for sparsity $s \in \{0.99, 0.9995\}$.

\looseness=-10000
\paragraph{\acronym{}.} A compressed binary mask over the target weights:
\looseness=-10000
\[
    \text{size}(\text{\acronym{}}) = \text{zstd}\bigl(\text{packbits}(\mathbf{1}[G > 0])\bigr).
\]
\looseness=-10000
No floating-point values are stored: kept weights are read from the public checkpoint, and zeroed weights are zero. At the learned sparsity below $0.05\%$ across the six GRPO runs in Table~\ref{tab:perf-combined}, zstd compresses to roughly $0.03$ bits per masked weight.

\section{Sparsity Learned by \acronym{} per Run}
\label{app:per-cell-sparsity}

\looseness=-10000
Table~\ref{tab:per-cell-sparsity} reports the fraction of all model weights that \acronymit{} deletes in the GRPO runs of Table~\ref{tab:perf-combined} ($\eta = 1/2048$, $\lambda = 10^{-5}$, $\mathit{lr} = 3{\times}10^{-4}$), at the earliest and final saved checkpoints. Every value stays below $0.05\%$.

\begin{table}[H]
\caption{
\looseness=-10000
\textbf{Fraction of all weights deleted by \acronymit{} in the six GRPO runs.} Fraction of all model parameters set to zero by the trained mask, at the earliest and final saved checkpoint. Every value is below $0.05\%$.
}
\label{tab:per-cell-sparsity}
\centering
\setlength{\tabcolsep}{6pt}
\small
\begin{tabular}{l rrr rrr}
\toprule
                & \multicolumn{3}{c}{\textbf{Llama, GSM8K}} & \multicolumn{3}{c}{\textbf{Qwen, SimpleRL-Zoo}} \\
\cmidrule(lr){2-4}\cmidrule(lr){5-7}
                & \textbf{1B} & \textbf{3B} & \textbf{8B} & \textbf{0.5B} & \textbf{3B} & \textbf{7B} \\
\midrule
Earliest saved  & $0.0342\%$ & $0.0420\%$ & $0.0456\%$ & $0.0393\%$ & $0.0379\%$ & $0.0381\%$ \\
Final saved     & $0.0347\%$ & $0.0422\%$ & $0.0459\%$ & $0.0405\%$ & $0.0418\%$ & $0.0415\%$ \\
\bottomrule
\end{tabular}
\end{table}

\section{Per-Benchmark Qwen2.5 Generalization Results}
\label{app:qwen-perbench}

\looseness=-10000
Table~\ref{tab:perf-qwen-perbench} reports the full per-benchmark breakdown that underlies the SimpleRL-Zoo (Qwen) accuracy columns of Table~\ref{tab:perf-combined}.

\begin{table}[H]
\caption{
\looseness=-10000
\textbf{Qwen2.5 on SimpleRL-Zoo: per-benchmark validation accuracy
(\%) at the best checkpoint.}
Six benchmarks: GSM8K, MATH500, Minerva Math, OlympiadBench, AIME24, AMC23.
}
\label{tab:perf-qwen-perbench}
\centering
\setlength{\tabcolsep}{4pt}
\small
\begin{tabular}{l rrrrrr}
\toprule
\textbf{Method} & GSM8K & MATH500 & Minerva & OlympiadBench & AIME24 & AMC23 \\
\midrule
\multicolumn{7}{l}{\textit{Qwen0.5B}} \\
FFT             & 48.07 & 36.00   &  4.78   &  7.73  &  0.00  & 17.50 \\
LoRA r=1        & 49.20 & 33.80   &  7.35   &  8.92  &  0.00  & 22.50 \\
LoRA r=4        & 52.46 & 36.60   &  8.82   & 10.55  &  3.33  & 10.00 \\
MTH 99\%        & 50.34 & 35.20   & 10.29   &  9.36  &  3.33  & 22.50 \\
MTH 99.95\%     & 49.73 & 35.20   &  6.25   &  7.28  &  0.00  & 22.50 \\
\rowcolor{TableHighlight}
\acronymit{}    & 44.12 & 33.40   &  5.51   &  6.98  &  3.33  & 12.50 \\
\midrule
\multicolumn{7}{l}{\textit{Qwen3B}} \\
FFT             & 83.62 & 66.20   & 23.16   & 26.45  & 16.67  & 50.00 \\
LoRA r=1        & 86.81 & 69.80   & 21.69   & 31.35  &  6.67  & 45.00 \\
LoRA r=4        & 85.60 & 67.00   & 22.79   & 28.68  & 10.00  & 47.50 \\
MTH 99\%        & 86.28 & 67.40   & 22.43   & 29.68  &  6.67  & 52.50 \\
MTH 99.95\%     & 85.37 & 63.80   & 22.79   & 27.93  & 13.33  & 50.00 \\
\rowcolor{TableHighlight}
\acronymit{}    & 83.78 & 66.20   & 22.79   & 25.78  & 13.33  & 40.00 \\
\midrule
\multicolumn{7}{l}{\textit{Qwen7B}} \\
FFT             & 91.43 & 78.20   & 33.46   & 39.92  & 13.33  & 55.00 \\
LoRA r=1        & 92.65 & 77.60   & 27.21   & 39.52  & 13.33  & 62.50 \\
LoRA r=4        & 91.66 & 77.00   & 29.04   & 39.97  & 20.00  & 52.50 \\
MTH 99\%        & 91.21 & 77.20   & 29.78   & 39.82  & 16.67  & 57.50 \\
MTH 99.95\%     & 92.04 & 75.00   & 31.62   & 38.78  & 20.00  & 50.00 \\
\rowcolor{TableHighlight}
\acronymit{}    & 91.28 & 78.80   & 32.72   & 40.29  & 23.33  & 47.50 \\
\bottomrule
\end{tabular}
\end{table}

\section{SFT Data Splits}
\label{app:sft-splits}

\looseness=-10000
For both MMLU-Pro and BBEH, we split each task's released set $75\%/25\%$ into train and held-out evaluation sets. Tables~\ref{tab:mmlu-pro-split} and~\ref{tab:bbeh-split} list per-task counts.

\begin{table}[H]
\caption{
\looseness=-10000
\textbf{Per-task train/test split for MMLU-Pro} under the $75/25$ split. Total: $9{,}018$ train / $3{,}014$ test across $14$ tasks.
}
\label{tab:mmlu-pro-split}
\centering
\small
\begin{tabular}{lrr@{\hskip 1.5em}lrr}
\toprule
Category & Train & Test & Category & Train & Test \\
\midrule
biology          & 537  & 180 & law         & 825  & 276 \\
business         & 591  & 198 & math        & 1013 & 338 \\
chemistry        & 849  & 283 & other       & 693  & 231 \\
computer science & 307  & 103 & philosophy  & 374  & 125 \\
economics        & 633  & 211 & physics     & 974  & 325 \\
engineering      & 726  & 243 & psychology  & 598  & 200 \\
health           & 613  & 205 & history     & 285  & 96  \\
\bottomrule
\end{tabular}
\end{table}

\begin{table}[H]
\caption{
\looseness=-10000
\textbf{Per-task train/test split for BBEH} under the $75/25$ split. Twenty-two tasks share the $200$-example pattern. The only outlier is \texttt{disambiguation\_qa} with $120$ examples. Total: $3{,}390$ train / $1{,}130$ test across $23$ tasks.
}
\label{tab:bbeh-split}
\centering
\small
\begin{tabular}{lrr}
\toprule
Task group & Train & Test \\
\midrule
$22$ tasks ($200$ examples each) & $150$ each & $50$ each \\
\texttt{disambiguation\_qa} ($120$ examples) & $90$ & $30$ \\
\bottomrule
\end{tabular}
\end{table}

\section{Full Wanda Sparsity Sweep}
\label{app:wanda-sweep}

\looseness=-10000
Tables~\ref{tab:wanda-sweep-llama} and~\ref{tab:wanda-sweep-qwen} report the per-sparsity breakdown underlying Section~\ref{sec:wanda-vs-base}'s best-Wanda picks. $\eta$ is the fraction of gate-projection weights deleted (same convention as Eq.~\eqref{eq:wanda-init}), and the sweep runs from $\eta = 1/2$ (half deleted) to $\eta = 1/8192$ ($\approx 0.012\%$ deleted). Wanda uses noise-free gate init ($G_{ij} = \pm 0.03$, hard-mask sparsity exactly $\eta$), while the trained \acronymit{} runs of Table~\ref{tab:perf-combined} use the noisy init of Eq.~\eqref{eq:wanda-init}. Qwen values are the six-benchmark mean across GSM8K, MATH500, Minerva Math, OlympiadBench, AIME24, and AMC23. ``Base'' is the model before fine-tuning. Bold marks the best Wanda accuracy per backbone.

\begin{table}[H]
\begin{minipage}[t]{0.46\textwidth}
\centering
\caption{\looseness=-10000 \textbf{Wanda sweep on Llama (\%).}}
\label{tab:wanda-sweep-llama}
\setlength{\tabcolsep}{4pt}
\small
\begin{tabular}{l rrr}
\toprule
$\eta$           & \textbf{Llama1B} & \textbf{Llama3B} & \textbf{Llama8B} \\
\midrule
$1/2$            &  0.00                 & 48.07                 & \textbf{60.65}      \\
$1/8$            &  1.36                 & 51.71                 & 59.14               \\
$1/32$           &  1.52                 & 51.55                 & 58.91               \\
$1/128$          &  1.52                 & 52.08                 & 59.67               \\
$1/512$          &  1.44                 & 52.16                 & 60.05               \\
$1/2048$         & \textbf{1.59}         & 52.16                 & 59.74               \\
$1/4096$         & \textbf{1.59}         & 51.63                 & 60.05               \\
$1/8192$         &  1.36                 & \textbf{52.46}        & 59.67               \\
\midrule
Base               &  1.21                 & 53.53                 & 60.27               \\
\bottomrule
\end{tabular}
\end{minipage}\hfill
\begin{minipage}[t]{0.46\textwidth}
\centering
\caption{\looseness=-10000 \textbf{Wanda sweep on Qwen (\%).}}
\label{tab:wanda-sweep-qwen}
\setlength{\tabcolsep}{4pt}
\small
\begin{tabular}{l rrr}
\toprule
$\eta$           & \textbf{Qwen0.5B} & \textbf{Qwen3B} & \textbf{Qwen7B} \\
\midrule
$1/2$            &  1.92                 & 21.06               & 39.39               \\
$1/8$            & 10.45                 & 36.09               & \textbf{42.49}      \\
$1/32$           & 11.24                 & 38.27               & 41.81               \\
$1/128$          & 11.21                 & 37.71               & 41.90               \\
$1/512$          & 10.56                 & 37.27               & 40.56               \\
$1/2048$         & \textbf{11.42}        & 38.24               & 40.14               \\
$1/4096$         & 11.09                 & 38.60               & 40.82               \\
$1/8192$         & 10.74                 & \textbf{38.70}      & 41.43               \\
\midrule
Base               & 10.44                 & 37.73               & 40.18               \\
\bottomrule
\end{tabular}
\end{minipage}
\end{table}

\section{Algebraic Connection to MaskLLM}
\label{app:maskllm-reduction}

\looseness=-10000
\acronymit{} and MaskLLM~\citep{maskllm} arrive at their forward passes from different starting points: \acronymit{} from a continuous relaxation of a per-weight binary masking decision (Eq.~\eqref{eq:train-fwd}), and MaskLLM from a Gumbel-Softmax over $N{:}M$ candidate patterns (Eq.~\eqref{eq:maskllm-soft}). After three structural modifications to Eq.~\eqref{eq:maskllm-soft} (binary candidates, no per-group cardinality, no Gumbel noise), the modified MaskLLM forward matches Eq.~\eqref{eq:train-fwd} algebraically.

\looseness=-10000
\paragraph{MaskLLM's forward.} The base weight matrix is partitioned into groups of $M$ contiguous weights, and each group is handled independently. Within a single group, MaskLLM enumerates $C(M,N) = \binom{M}{N}$ candidate binary patterns $\widehat{M}_c$ with $c \in \{0,\dots,C(M,N)-1\}$, each keeping exactly $N$ of $M$ entries. It places a learnable logit vector $\pi \in \mathbb{R}^{C(M,N)}$ over the candidates and draws a Gumbel-Softmax sample~\citep{gumbelsoftmax,concrete}:
\begin{equation}
\widetilde{y}_c \;=\; \frac{\exp\!\big((\kappa\,\pi_c + g_c)/\tau\big)}{\sum_{c'} \exp\!\big((\kappa\,\pi_{c'} + g_{c'})/\tau\big)},
\qquad g_c \stackrel{\mathrm{iid}}{\sim} \mathrm{Gumbel}(0,1).
\label{eq:maskllm-soft}
\end{equation}
\looseness=-10000
The group's soft mask is the candidate-pattern-weighted average $\widetilde{M} = \sum_c \widetilde{y}_c\widehat{M}_c$, and the layer forward applies $W \odot \widetilde{M}$ within the group. At evaluation, the hard mask is $\widehat{M}_{c^\star}$ with $c^\star = \arg\max_c \pi_c$. The temperature $\tau$ and logit-scale $\kappa$ both follow linear schedules over training.

\looseness=-10000
\paragraph{Step 1: $C(M,N){=}2$ collapses the two-candidate softmax to a sigmoid.} Set $M{=}2, N{=}1$, so each group has exactly two candidates: $\widehat{M}_0 = (1,0)$ keeps position 0 and $\widehat{M}_1 = (0,1)$ keeps position 1. With only two candidates, the Gumbel-Softmax probability on candidate 0 simplifies to a sigmoid of the logit difference:
\begin{equation}
\widetilde{y}_0
= \frac{e^{(\kappa\pi_0 + g_0)/\tau}}{e^{(\kappa\pi_0 + g_0)/\tau} + e^{(\kappa\pi_1 + g_1)/\tau}}
= \frac{1}{1 + e^{-\frac{1}{\tau}[\kappa(\pi_0-\pi_1) + (g_0-g_1)]}}
= \sigma\!\Big(\tfrac{1}{\tau}\big[\kappa\,(\pi_0 - \pi_1) + (g_0 - g_1)\big]\Big),
\end{equation}
\looseness=-10000
where the second equality divides numerator and denominator by the first exponential, and the third uses $\sigma(x) = 1/(1+e^{-x})$.
The difference of two independent $\mathrm{Gumbel}(0,1)$ variables is $\mathrm{Logistic}(0,1)$, so
\begin{equation}
\widetilde{y}_0 \;=\; \sigma\!\big(\tfrac{1}{\tau}[\kappa\,\Delta\pi + \xi]\big),
\qquad \Delta\pi := \pi_0 - \pi_1,\ \ \xi \sim \mathrm{Logistic}(0,1).
\label{eq:binary-concrete}
\end{equation}
\looseness=-10000
Eq.~\eqref{eq:binary-concrete} is the Binary-Concrete distribution~\citep{concrete}.

\looseness=-10000
\paragraph{Step 2: drop the per-group cardinality constraint.} MaskLLM forces each group of $M$ weights to keep exactly $N$ of them. Letting each weight choose its own binary mask value replaces this group categorical with an independent binary masking choice per weight, parameterized by a single logit $G_{ij}$. Each weight then has its own Binary-Concrete (Eq.~\eqref{eq:binary-concrete} applied per-weight, with $G_{ij}$ in place of $\Delta\pi$):
\begin{equation}
\widetilde{M}_{ij} \;=\; \sigma\!\big(\tfrac{1}{\tau}(\kappa\,G_{ij} + \xi_{ij})\big), \qquad \xi_{ij} \sim \mathrm{Logistic}(0,1).
\end{equation}
\looseness=-10000
\paragraph{Step 3: drop the noise.} Setting the noise $\xi_{ij}$ to zero in Step 2's per-weight Binary-Concrete gives a deterministic soft mask:
\begin{equation}
\widetilde{M}_{ij} \;=\; \sigma\!\big(\tfrac{\kappa}{\tau}\,G_{ij}\big).
\end{equation}
\looseness=-10000
This matches Eq.~\eqref{eq:train-fwd} with the scalar $\alpha(t) := \kappa(t)/\tau(t)$; in our implementation, we schedule this scalar directly and linearly.

\section{\acronym{} Hyperparameter Sweeps}
\label{app:hp-sweep}

\looseness=-10000
Tables~\ref{tab:hp-sweep-llama} and~\ref{tab:hp-sweep-qwen} give every setting of the sweep in Section~\ref{sec:hp-sensitivity}.

\begin{table}[H]
\begin{minipage}[t]{0.483\textwidth}
\caption{
\looseness=-10000
\textbf{Robustness to hyperparameters on Llama8B GSM8K.} Best-checkpoint accuracy (\%) of \acronymit{} for each $(\eta, \lambda, \mathit{lr})$.
}
\label{tab:hp-sweep-llama}
\centering
\setlength{\tabcolsep}{4pt}
\small
\begin{tabular}{c rr@{\hspace{18pt}}rr}
\toprule
& \multicolumn{2}{c}{$\lambda = 0$}
& \multicolumn{2}{c}{$\lambda = 10^{-5}$} \\
\cmidrule(lr){2-3}\cmidrule(lr){4-5}
$\eta\,\backslash\,\mathit{lr}$ & $10^{-4}$ & $3{\times}10^{-4}$ & $10^{-4}$ & $3{\times}10^{-4}$ \\
\midrule
$1/2048$ & 83.40        & 82.49                 & 82.34        & \textbf{83.55}        \\
$1/4096$ & 83.02        & 82.03                 & 83.09        & 83.09                 \\
$1/8192$ & 83.40        & 82.49                 & 83.09        & 83.02                 \\
\bottomrule
\end{tabular}
\end{minipage}
\hfill
\begin{minipage}[t]{0.483\textwidth}
\caption{
\looseness=-10000
\textbf{Robustness to hyperparameters on Qwen7B SimpleRL-Zoo.} Best-checkpoint accuracy (\%) of \acronymit{} for each $(\eta, \lambda, \mathit{lr})$.
}
\label{tab:hp-sweep-qwen}
\centering
\setlength{\tabcolsep}{4pt}
\small
\begin{tabular}{c rr@{\hspace{18pt}}rr}
\toprule
& \multicolumn{2}{c}{$\lambda = 0$}
& \multicolumn{2}{c}{$\lambda = 10^{-5}$} \\
\cmidrule(lr){2-3}\cmidrule(lr){4-5}
$\eta\,\backslash\,\mathit{lr}$ & $10^{-4}$ & $3{\times}10^{-4}$ & $10^{-4}$ & $3{\times}10^{-4}$ \\
\midrule
$1/2048$ & 51.79        & 51.92                 & 49.60        & 52.32                 \\
$1/4096$ & 52.23        & \textbf{52.80}        & 50.29        & 51.91                 \\
$1/8192$ & 51.14        & 52.10                 & 51.29        & 52.34                 \\
\bottomrule
\end{tabular}
\end{minipage}
\end{table}

\section{FFT Hyperparameter Sweeps}
\label{app:fft-sweep}
\label{app:sft-fft-sweep}

\looseness=-10000
To make sure the FFT baseline is not under-tuned, we sweep its learning rate under GRPO and SFT and its KL coefficient under GRPO. All other settings follow the main runs. None of these settings changes the comparison with \acronymit{} (Tables~\ref{tab:grpo-fft-sweep} and~\ref{tab:sft-fft-sweep}).

\begin{table}[H]
\caption{\looseness=-10000 \textbf{FFT sweep under GRPO.} Best-checkpoint validation accuracy (\%), as in Table~\ref{tab:perf-combined}.}
\label{tab:grpo-fft-sweep}
\centering
\small
\begin{tabular}{l r r r r}
\toprule
\textbf{Method} & \textbf{Learning rate} & \textbf{KL coefficient} & \textbf{Llama8B, GSM8K} & \textbf{Qwen7B, SimpleRL-Zoo} \\
\midrule
FFT & $5{\times}10^{-7}$ & $0$ & 83.23 & 50.55 \\
FFT & $5{\times}10^{-7}$ & $10^{-4}$ & 82.55 & 51.89 \\
FFT & $10^{-6}$ & $0$ & 83.24 & 51.93 \\
FFT & $10^{-6}$ & $10^{-4}$ & 81.89 & 50.79 \\
\rowcolor{TableHighlight}
\acronymit{} & $3{\times}10^{-4}$ & as in Table~\ref{tab:perf-combined} & \textbf{83.55} & \textbf{52.32} \\
\bottomrule
\end{tabular}
\end{table}

\begin{table}[H]
\caption{\looseness=-10000 \textbf{FFT learning-rate sweep under SFT, Llama3B MMLU-Pro.} Best-checkpoint accuracy (\%), as in Table~\ref{tab:perf-sft}.}
\label{tab:sft-fft-sweep}
\centering
\small
\begin{tabular}{l r r}
\toprule
\textbf{Method} & \textbf{Learning rate} & \textbf{Accuracy} \\
\midrule
Base & -- & 12.63 \\
FFT & $10^{-5}$ & 19.35 \\
FFT & $3{\times}10^{-6}$ & 21.19 \\
FFT & $10^{-6}$ & 18.35 \\
\rowcolor{TableHighlight}
\acronymit{} & $5{\times}10^{-4}$ & \textbf{22.28} \\
\bottomrule
\end{tabular}
\end{table}

\newpage

\end{document}